\documentclass[10pt,twocolumn,letterpaper]{article}

\usepackage{iccv}
\usepackage{times}
\usepackage{epsfig}
\usepackage{graphicx}
\usepackage{caption}
\usepackage{capt-of}
\usepackage{amsmath,mathtools}
\usepackage{amssymb}
\usepackage{algorithm} 
\usepackage{algpseudocode}
\usepackage{booktabs}
\usepackage[tight]{subfigure}
\usepackage{array}
\usepackage{breqn}
\usepackage{bm}

\usepackage[pagebackref=true,breaklinks=true,letterpaper=true,colorlinks,bookmarks=false]{hyperref}
\DeclareCaptionFont{small}{\small}
\iccvfinalcopy 

\def\iccvPaperID{79} 

\ificcvfinal\fi
\begin{document}

\renewcommand{\floatpagefraction}{.8}
\renewcommand{\topfraction}{.75}
\newcolumntype{C}{>{\centering\arraybackslash}p{4.165em}}
\newcolumntype{D}{>{\centering\arraybackslash}p{3.95em}}

\subfigbottomskip = -1cm
\setlength{\parskip}{0cm}

\newcommand{\bld}[1]{{\bf{#1}}}

\newcommand{\horrule}[1]{\rule{\linewidth}{#1}}     

\newcommand{\insertimage}[4]{ 
\begin{figure}[t]
\centering
\includegraphics[scale=#1, clip=true]{figures/#2}
\caption{#3}
\label{#4}
\end{figure}
}

\newcommand{\insertimageC}[5]{ 
\begin{figure}[#5]
\centering
\includegraphics[width=#1\linewidth, clip=true]{figures/#2}
\caption{#3}
\vspace{-0.5em}
\label{#4}
\end{figure}
}

\newcommand{\insertimageStar}[5]{ 
\begin{figure*}[#5]
\centering
\includegraphics[width=#1\linewidth, clip=true]{figures/#2}
\caption{#3}
\vspace{-0.5em}
\label{#4}
\end{figure*}
}

\algnewcommand\algorithmicinput{\textbf{Input:}}
\algnewcommand\INPUT{\item[\algorithmicinput]}
\algnewcommand\algorithmicoutput{\textbf{Output:}}
\algnewcommand\OUTPUT{\item[\algorithmicinput]}

\DeclarePairedDelimiter\ceil{\lceil}{\rceil}
\DeclarePairedDelimiter\floor{\lfloor}{\rfloor}

\newcommand{\cSlo}[1]{\textcolor{red}{#1}}
\newcommand{\cTolga}[1]{\textcolor{blue}{#1}}
\newcommand{\comment}[1]{}

\newcommand{\argmin}{\operatornamewithlimits{argmin}}
\newcommand{\argmax}{\operatornamewithlimits{argmax}}
\let\oldemptyset\emptyset
\let\emptyset\varnothing
\newcommand*{\argminl}{\argmin\limits}
\newcommand*{\argmaxl}{\argmax\limits}

\setlength{\textfloatsep}{10pt}

\title{CAD Priors for Accurate and Flexible Instance Reconstruction}

\author{Tolga Birdal$^{1,2}$ \qquad\qquad Slobodan Ilic$^{1,2}$\\
$^1$ Technische Universit{\"a}t M{\"u}nchen, Germany\\%
$^2$ Siemens AG, Munich, Germany\\%
}

\twocolumn[{
\renewcommand\twocolumn[1][]{#1}%
\maketitle
\vspace{-32pt}
\begin{center}
    \centering
    \includegraphics[width=\textwidth]{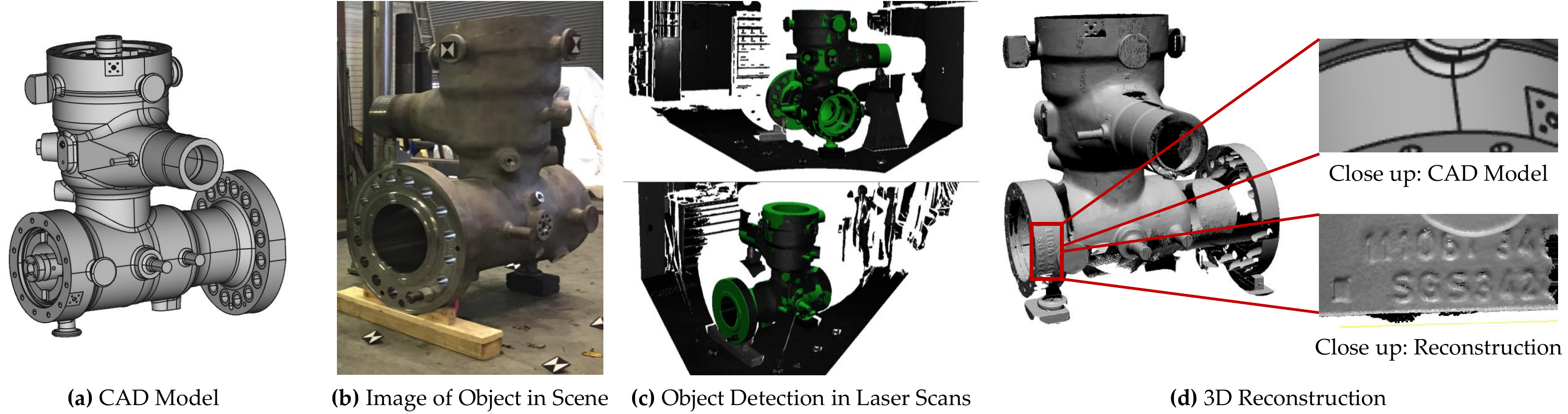}
    \captionof{figure} {\small{Our 3D reconstruction method. \textit{(a)} Input 3D CAD model. \textit{(b)} Image of the instance to reconstruct. \textit{(c)} Detection of 3D model in point clouds. \textit{(c)} Final reconstruction we obtain, with close-up comparisons to the nominal CAD prior.}}
    \label{fig:teaser}
\end{center}%
}]
\begin{abstract}
We present an efficient and automatic approach for accurate instance reconstruction of big 3D objects from multiple, unorganized and unstructured point clouds, in presence of dynamic clutter and occlusions. In contrast to conventional scanning, where the background is assumed to be rather static, we aim at handling dynamic clutter where the background drastically changes during object scanning. Currently, it is tedious to solve this problem with available methods unless the object of interest is first segmented out from the rest of the scene. We address the problem by assuming the availability of a prior CAD model, roughly resembling the object to be reconstructed. This assumption almost always holds in applications such as industrial inspection or reverse engineering. With aid of this prior acting as a proxy, we propose a fully enhanced pipeline, capable of automatically detecting and segmenting the object of interest from scenes and creating a pose graph, online, with linear complexity. This allows initial scan alignment to the CAD model space, which is then refined without the CAD constraint to fully recover a high fidelity 3D reconstruction, accurate up to the sensor noise level. We also contribute a novel object detection method, local implicit shape models (LISM) and give a fast verification scheme. We evaluate our method on multiple datasets, demonstrating the ability to accurately reconstruct objects from small sizes up to $125m^3$.
\end{abstract}

\section{Introduction}
3D reconstruction involves the recovery of digitized object or scene using a 2D/3D sensor, typically through multiple acquisition steps. From reverse engineering to industrial inspection, its applications are plentiful. Due to such wide use, from the early days of vision, it attracted significant attention of both research community and industry~\cite{Levoy:2000:TDM,blais1995}.

Despite the huge demand, many marker-free approaches based solely on 3D data either involve acquisition of ordered scans
\cite{newcombe2011kinectfusion,kehl2014bmvc}, or follow the de-facto standard pipeline \cite{huber2003fully} in case of unordered scans. The former suffers from the requirements of redundant depth capture with large overlap and scenes with very little clutter or occlusions. Due to the volumetric nature of scan fusion, such techniques also do not scale well to large objects while retaining high precision. The latter exploits 3D keypoint matching of all scans to one another, alleviating the order constraint. Thanks to 3D descriptors, it could well operate in full 3D. Yet, matching of scans to each other is an $O(N^2)$ problem and prevents the methods from scaling to an arbitrary number of scans. In addition, neither of those can handle scenes with extensive dynamic clutter or occlusions.

Nowadays, with the capability of collecting high quality, large scale and big data, it is critical to offer automated solutions for providing highly accurate reconstructions regardless of the acquisition scenarios. In this paper, we tackle the problem of 3D instance reconstruction from a handful of unorganized point clouds, where the object of interest is large in size, texture-less, surrounded by significant dynamic background clutter and is viewed under occlusions. Our method can handle scenes between which no single transformation exists, i.e. the same objects appear in different locations, such as the one in Fig. \ref{fig:dynamic}. We also do not impose any constraints on the order of acquisition. To solve all of these problems simultaneously, we make use of the reasonable assumption that a rough, nominal 3D CAD model prior of the object-to-reconstruct is available beforehand and propose a novel reconstruction pipeline. 
Such assumption of a nominal prior is valid for many applications especially in industry, where the objective is to compare the reconstruction to the designed CAD model. Even for the cases where this model does not exist, one could always generate a rough, inaccurate mesh model with existing methods, e.g. KinectFusion \cite{newcombe2011kinectfusion}, to act as a prior. Note that, due to manufacturing errors, sensor noise, damages and environmental factors, physical instances deviate significantly from the CAD models and the end-goal is an automatic algorithm to accurately recover the particular instance of the model.
With the introduction of the prior model, we re-factor the standard 3D reconstruction procedures via multi-fold contributions. We replace the scan-to-scan matching with model-to-scan matching resulting in absolute poses for each camera. Unlike the case in object instance detection where false positives (FP) are tolerable, object instance reconstruction is easily jeopardized by the inclusion of a single FP. Therefore, one of our goals is to suppress FP, even at the expense of some true negatives. To achieve this, we contribute a probabilistic Local Implicit Shape Model (LISM) formulation for the object instance detection and pose estimation, accompanied by a rigorous hypotheses verification to reject all wrong pose candidates. This matching is followed by automatically segmenting out the points belonging to the object and transforming them back onto the model coordinate frame. Doing this for multiple views results in roughly aligned partial scans in the CAD space. To fully recover the exact object, the CAD prior is then discarded (as it might cause undesired bias) and a global multi-view refinement is conducted only to optimize the camera poses. 

In global scan registration, the creation of a \textit{pose graph} indicating which scans can be registered to each other is required. The typical complexity of obtaining this graph is $O(N^2)$, where all scans are matched to one another. We profit from the CAD prior and contribute by automatically computing this graph, in which only cameras sharing significant view overlap are linked. This reduces the complexity to $O(N)$, and robustifies the whole pipeline.

The entire procedure is made efficient so that large scans are handled in reasonable time. Exhaustive evaluations demonstrate high accuracy regardless of the size of the objects, clutter and occlusions.\footnote{Our suppl. video is under: \url{https://youtu.be/KPA_8BNuOvg}}
\insertimageC{1}{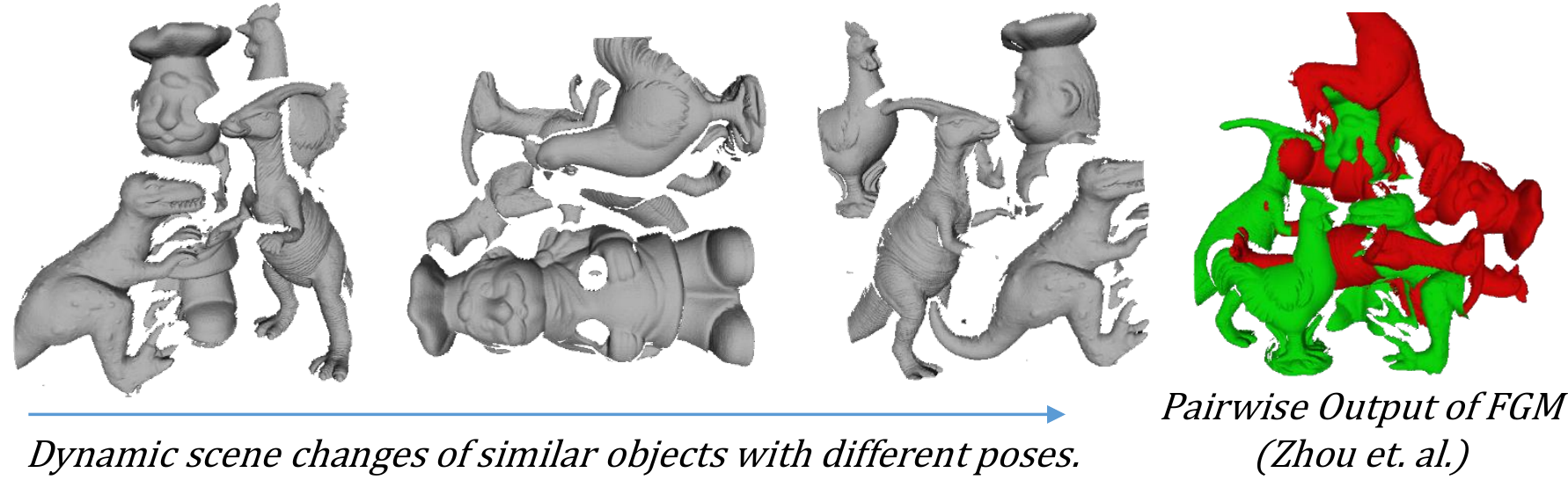}{\small{Dynamic Clutter and Occlusions on Mian Dataset \cite{mian2006three}: There is change in relative locations of the objects between scenes. This easily fools the modern global registration algorithms \cite{zhou2016fast}. We operate on the object level and circumvents this problem.}}{fig:dynamic}{t!}

\section{Prior Art}
Arguably, the most wide-spread 3D reconstruction methods are KinectFusion\cite{newcombe2011kinectfusion} and its derivatives \cite{canelhas2013sdf,kehl2014bmvc,slavcheva2016eccv}. These methods have been successful in reconstructing small isolated objects, but their application is not immediate when the size increases, or clutter and occlusions are introduced \cite{niessner2013hashing}. Due to extensive usage of signed-distance fields, they are bound to depth images and a sequential acquisition. 

Abundance of works exists in multi-view global alignment from multiple 3D unorganized point clouds \cite{arrigoni2016global,danelljan2016probabilistic,evangelidisH16,eckartaccelerated,torsello2011multiview,zhu2014robust,Tang2015,govindu2014averaging,toldo2010global,krishnan2005global,bonarrigo2011enhanced,eckartaccelerated,zhou2016fast}. These methods assume the scans to be roughly initialized and reasonably well-segmented. They, in general, handle slight synthetic noise well enough, but they do not deal with cluttered and occluded data. 
Another track of works try to overcome the aforementioned constraints by using keypoint detectors and matching descriptors in 3D scans. These methods operate on a subset of points during matching. One of the pioneering works proposing a feature agnostic, automatic and constraint-free algorithm is the graph based in-hand scanning from D. Huber and M. Hebert~\cite{huber2003fully}. The authors set a baseline for this family of methods. Novatnack and Nishino \cite{novatnack2008scale} developed a scale dependent descriptor for better initialization and fused it with \cite{huber2003fully} to assess the power of their descriptor. Yet both of these relied on range image data. Mian et. al.~\cite{mian2006three} proposed a tensor feature and a hashing framework operating on meshes. Fantoni et al.~\cite{fantoni2012accurate} uses 3D keypoint matching as an initial stage of multiview alignment to bring the scans to a rough alignment. Zhu et. al. \cite{zhu2014robust,zhu2016automatic} as well as Liu and Yonghuai \cite{liu2005automatic} use genetic algorithms to discover the matching scans and use this in global alignment context. These stochastic schemes are correspondence free but slow. Similar to ~\cite{mian2006three}, Zhu et. al. \cite{zhu2016local} devises a local-to-global minimum spanning tree method to align the scans. A majority of these automatic alignment procedures suffer from increased worst case complexity of $O(N^2)$, where $N$ is the number of views. Moreover, since there is no integrated segmentation, the registration procedure cannot handle clutter and occlusions.

Use of CAD models in reconstruction is not novel by itself. Savarese et. al. \cite{yingze2013dense} enrich the multiview reconstruction from 2D images with a CAD prior. Guney and Geiger \cite{guney2015displets} use object knowledge to resolve the stereo ambiguities. Birdal et. al. use models in triangulation by registration \cite{birdal2016online}. Other works \cite{liebelt2010multi,yan20073d,zia2013detailed} use CAD prior to detect generic object classes. Our approach differs from all of those in the sense that we use proxy instance priors for initial alignment of scans and then operate directly on 3D points.  

Salas et. al. \cite{salas2013slampp} propose SLAM++ using object priors to constrain a SLAM system. Our work is differentiated from theirs in the sense that we perform instance reconstruction using the CAD prior, and not SLAM. In our setting, the background as well as the object is allowed to change dynamically between different acquisitions.

\insertimageC{0.9}{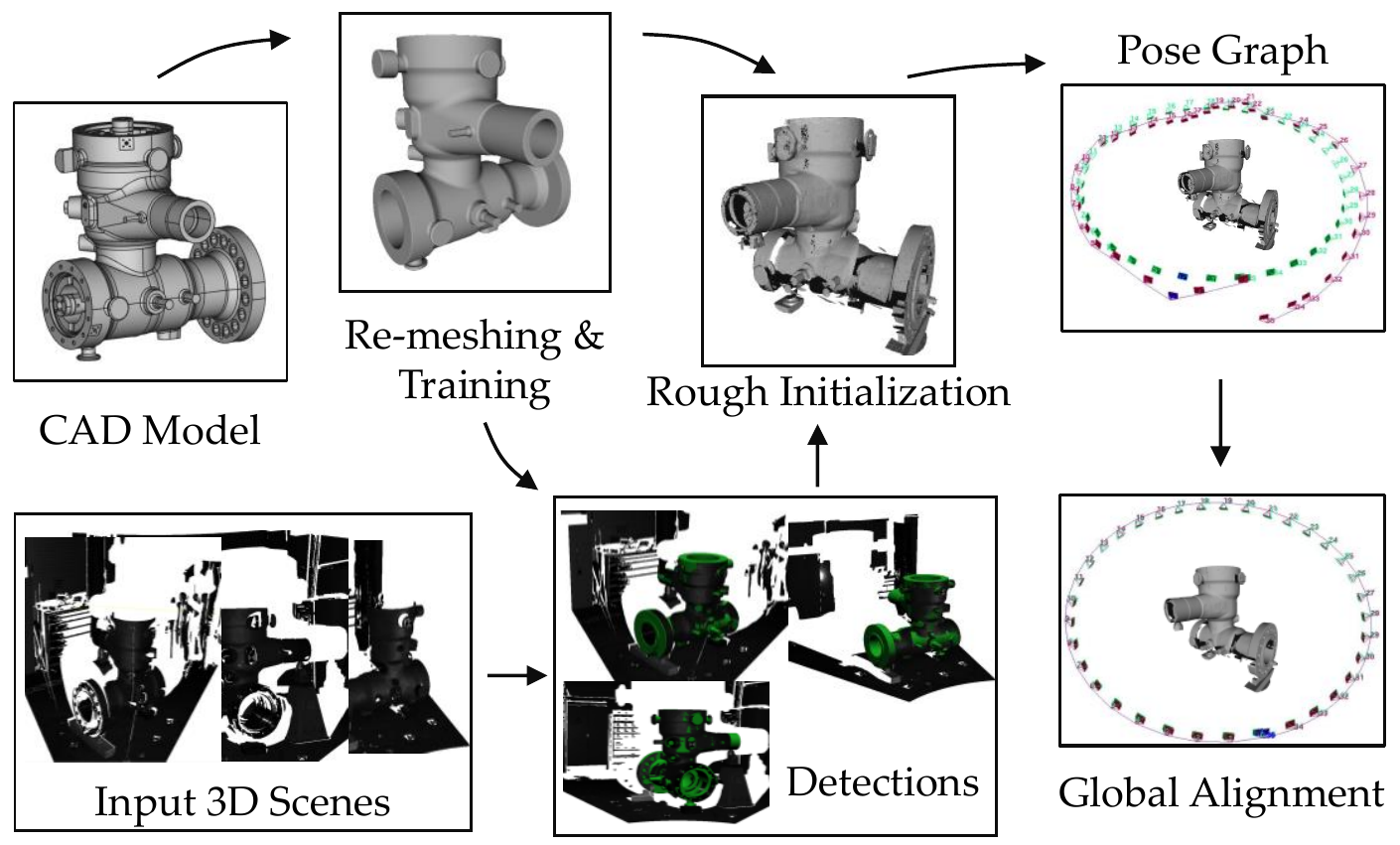}{\small{Proposed 3D reconstruction pipeline: Prior CAD model is trained to create the model representation. Input scenes are then parsed for the model pose. Pose estimates initialize a rough reconstruction, with segmentation and automatically computed pose graph. This is then further refined to the full reconstruction.}}{fig:pipeline}{t!}

\section{Method}
\label{sec:method}
Given a set of unstructured and unordered 3D scenes $\{\mathbf{S}_i\} \in \mathbf{S}$, we seek to find a set of transformations $\{\mathbf{T}_i\} \in SE(3)$ so as to stitch and reconstruct a global model $\mathbf{S}_G$: 
\vspace{-1pt}
\begin{equation}
\label{eq:stitch}
\mathbf{S}_G = \bigcup_{i=1}^N \big(\mathbf{T}_i^0 \circ \mathbf{S}_i\big)
\end{equation}
$\mathbf{T} \circ \mathbf{S}$ applies transformation $\mathbf{T}$ to the scene $\mathbf{S}$. In this setting, both the transformations $\{\mathbf{T}_i^0\} \in SE(3)$ and the global model $\mathbf{S}_G$ are unknown and we do not assume known initialization. Due to the lack of a common reference frame and apriori information about $\{\mathbf{S}_i\}$, obtaining the set $\{\mathbf{T}_i^0\}$ typically requires $O(N^2)$ worst case complexity, where all the scene clouds are matched to one another to obtain the relative transformations aligning them. To better condition the problem and reduce its complexity, we introduce the supervision of a CAD proxy $\mathbf{M}$ in form of a mesh model and re-write Eq. \eqref{eq:stitch}:
\vspace{-3pt}
\begin{equation}
\mathbf{S}_G= \bigcup _{i=1}^N \Big(\mathbf{T}_i^M \circ (S_i | \mathbf{M})\Big)
\end{equation}
where $\mathbf{T}_i^M \in SE(3)$ is the transformation from the scene to the model space, such that the segmented scene points $(\mathbf{S}_i | \mathbf{M})$ come to the best agreement. To estimate $\{\mathbf{T}_i^M\}$, we follow a two stage technique. First, a rather approximate estimate $\{\mathbf{\tilde{T}}_i^M\}$ is made by matching the CAD model to a single scene. Note that this time, the set $\{\mathbf{\tilde{T}}_i^M\}$ can be computed in $O(N)$, since it only requires CAD to scan alignment. However, because scene clouds suffer from partial visibility, noise and deviations w.r.t. the CAD model, the discovery of the pose of the model in the scans provides only rough initial transformations to the model coordinate frame. For this reason, as the final stage, the CAD prior is disregarded and the scans are globally refined, simultaneously. This lets us reconstruct configurations deviating significantly from the CAD prior.


Our procedure of multi-view refinement is similar to ~\cite{fantoni2012accurate}, where a global scheme for scan alignment is employed. Let $\mathbf{S}_1, \dots \mathbf{S}_M$ be the set of scans that are to be brought in alignment. To generalize and formalize the notation 
for registrations of all point clouds to each other, we maintain a directed pose graph in form of an adjacency matrix $A \in \{0,1\}^{M \times M}$, such that $A(h,k) = 1$ iff cloud $\mathbf{S}_h$ can be registered to cloud $\mathbf{S}_k$. Let $\bm{\theta}=(\bm{\theta}_1, \dots, \bm{\theta}_M)$ be the absolute camera poses of each view. The alignment error between two clouds $\mathbf{S}_h$ and $\mathbf{S}_k$ then reads:
\begin{equation}
\label{eq:mvlmicp-pair}
E(\bm{\theta}_h, \bm{\theta}_k) =  A(h,k) \sum_{i=1}^{N_h} \rho \bigg( \lVert d(\bm{\theta}_h \circ p_i^h, \bm{\theta}_k \circ q_i^h) \rVert^2 \bigg) 
\end{equation}
where $\{p_i^h \rightarrow q_i^h\}$ are the $N_h$ closest point correspondences obtained from the clouds $\mathbf{S}_h$ and $\mathbf{S}_k$. The point-to-plane distance $d(\cdot, \cdot)$ is defined to be:
\begin{equation}
d(p_i, q_i) = (\mathbf{R} p_i + \mathbf{t}-q_i)^T n^q_i
\end{equation}
with $n^q_i$ referring to the normal associated to point $q_i$. $\mathbf{R}$ and $\mathbf{t}$ are the components of the pair transformation $\{\mathbf{R} \in SO(3), \mathbf{t} \in \mathbb{R}^3\}$.
The overall alignment error, which we want to minimize at this stage, is obtained by summing up the contribution of every pair of overlapping views:
\begin{equation}
\label{eq:mvlmicp-general}
E(\bm{\theta}) =\sum_{h=1}^M \sum_{k=1}^M A(h,k) \sum_{i=1}^{N_h} \rho \bigg(\lVert d(\bm{\theta}_h  \circ p_i^h, \bm{\theta}_k  \circ q_i^h) \rVert^2  \bigg)
\end{equation}
where $\rho$ is the robust estimator. The final absolute poses are the result of the minimization $(\bm{\theta}_1, ..., \bm{\theta}_M) = \argmin_{\bm{\theta}}(E)$, and align the $M$ clouds in a least squares sense. In contrast to the pairwise registration error in Eq. \eqref{eq:mvlmicp-pair}, which has closed form solution for the relative transformation $\bm{\theta}$, there are no closed form solutions in the multiview setting. Therefore, we use a non-linear optimization procedure, Levenberg Marquardt. The rotations are parameterized with angle-axis representation $(\mathbf{w} \in \mathbb{R}^3, \phi)$. We constrain the frame with the highest number of points found on the CAD model to be static and update (solve for) the rest of the poses. In practice, this leads to faster convergence. Note that, in contrast to the methods that exploit pairwise registration, our poses are absolute and do not suffer from drifts or tracking artifacts. We also do not require a conversion from relative poses to absolute ones, which are usually obtained by the computation of a minimum spanning tree or shortest paths over the pose graph \cite{govindu2006robustness,huber2003fully}. This property eases the implementation and reduces errors, that are to be encountered in usual heuristics.

Due to the accuracy requirements, unlike~\cite{fantoni2012accurate}, we omit using distance transforms at this stage. We rather use speeded up KD-Trees to achieve exact nearest neighbors~\cite{muja2014scalable}. Since we optimize over the poses, and not over 3D points, the trees are built only once in the beginning and all closest point computations are done in the local coordinate frame of the view of interest. This is important for efficiency. For reasons of accuracy, we use analytical Jacobians. As cloud sizes become large, this optimization exhibits significant computational costs. This is why, a priori sampling plays a huge role, where we use $\approx$ 20k to 30k points per scan, distributed evenly in space.

To summarize, our key contribution lies in obtaining $\{\mathbf{T}_i^M\}$ in a robust, efficient and accurate manner. We will now show how to compute the rough alignment $\{\mathbf{\tilde{T}}_i^M\}$ and the pose graph (adjacency matrix) $A$.

\insertimageC{1}{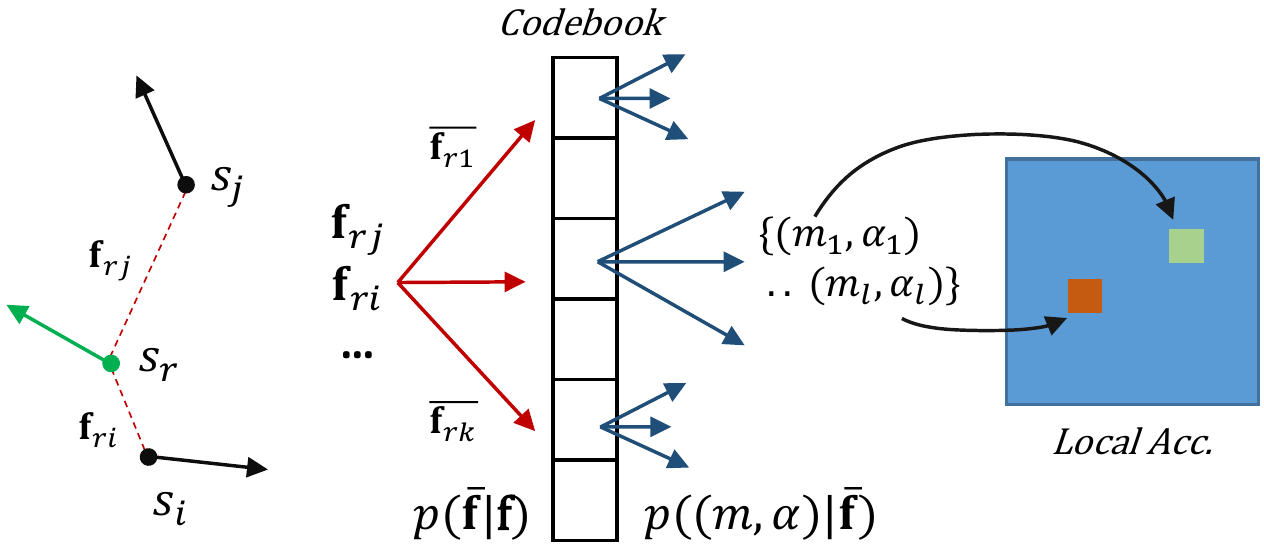}{\small{Local Implicit Voting: Given multiple scene point pairs, tied to a common reference $\mathbf{s}_r$, we generate features $\mathbf{f}_{ri}$, activating different codebook buckets (middle). Each bucket casts votes for multiple $(m,\alpha)$ pairs in the local voting space of $\mathbf{s}_r$.}}{fig:ism}{t!}

\subsection{Locally Implicit Models for Estimating $\{\mathbf{\tilde{T}}_i^M\}$}
While using any method, which is capable of handling 3D points, e.g.~\cite{johnson1999using,tombari2012hough,mian2006three,tombari2010unique,papazov2012rigid}, is possible, we mainly follow the Geometric Hashing of Drost et al.~\cite{drost2010model} and Birdal and Ilic~\cite{birdal20153dv} due to efficiency and robustness to clutter and occlusions. 
Yet, we introduce a more effective probabilistic formulation, inspired by the implicit shape models~\cite{Leibe04combinedobject}.


\paragraph{Model Description} 
In the first stage, we generate a pose invariant \textit{codebook} encoding all possible semi-global structures that could be found on the CAD model. We represent this semi-global geometry via simple point pair features (PPF) of oriented point pairs $(\mathbf{m}_i, \mathbf{m}_j)$:
\begin{equation}
\mathbf{f}_{ij} = (\Vert \mathbf{d}\rVert_2, \angle(\mathbf{n}_i,\mathbf{d}), \angle(\mathbf{n}_j,\mathbf{d}), \angle(\mathbf{n}_i,\mathbf{n}_j))
\end{equation}
where $\mathbf{d}=\mathbf{m}_i-\mathbf{m}_j$, $\mathbf{n}_i$ and $\mathbf{n}_j$ are the surface normals at points $\mathbf{m}_i$ and $\mathbf{m}_j$. $\angle(\cdot,\cdot)$ is the angle operator. The complete set of such features $\mathbf{F}=\{\mathbf{f}_{ij}\}$ for the prior CAD model is collected and quantized to generate the codebook: $\hat{F}$. We use our codebook to relate a feature $\mathbf{f}$ (key) to a set of oriented references points $\{(\mathbf{m}_i,\mathbf{m}_j)\}$ (stored in buckets) and build the global model description as an inverted file i.e. a hashtable $\mathbf{H}$. Thus, each bucket in the codebook contains self-similar point pairs extracted from the CAD model. Whenever a pair from the scene is matched to one in the model, their normals at the reference points are aligned. Then, the full pose of the object can be obtained once the rotation angle $\alpha$ around the normal is known. This can be done by aligning Local Coordinate Frames (LCF) constructed from matched pairs. Thus, instead of storing the full PPF, we store only this local parameterization $\{\mathbf{m}_r,\alpha \}$  composed of the model reference point $\mathbf{m}_r$ and rotation angle $\alpha$. A pair correspondence resolves the full 6DOF pose and what is left is to retrieve the matching pairs $\{\mathbf{s}_r, \mathbf{s}_i\}$ and $\{\mathbf{m}_r, \mathbf{m}_i\}$. We now give a novel way to do this.

\comment{\cSlo{
We now cast the detection and pose estimation problem as retrieving/matching the correspondence of an oriented scene point pair $\{\mathbf{s}_r, \mathbf{s}_i\}$ to the one on the model $\{\mathbf{m}_r, \mathbf{m}_i\}$. 
}}
\comment{
We now cast the detection and pose estimation problem as retrieving the correspondence of an oriented point pair $\{\mathbf{m}_r, \mathbf{m}_i\}$ to the scene analogous $\{\mathbf{s}_k, \mathbf{s}_j\}$. Instead of using the pairs directly, we parameterize it by a reference point $\mathbf{m}_r$ and the rotation of the paired point $\mathbf{m}_i$ around the normal of $\mathbf{m}_r$: $\alpha_{ri}$. Note that such pair correspondence fully resolves the 6DOF pose, once the relative rotations around normals are aligned: $\alpha=\alpha_s-\alpha_m$. Therefore, we use our codebook to relate a feature $\mathbf{f}$ (key) to a set of oriented references points $\{(\mathbf{m}_k,\alpha_k)\}$ (buckets) and build the global model description as an inverted file i.e. a hashtable $\mathbf{H}$.
} 

\paragraph{Probabilistic Formulation} During detection, a new point cloud scene $\mathbf{S}$ is encountered and downsampled to a set of points $\mathbf{S}_D=\{\mathbf{s}_r\}$, some of which are assumed to lie on the object. The sampling also enforces spatial uniformity (see our suppl. material). We fix a reference point $\mathbf{s}_r$ and pair it with all the other samples $\{\mathbf{s}_i\}$. Each pair makes up a PPF $\mathbf{f}_{ri}$. The original method~\cite{drost2010model} associates $\mathbf{f}_{ri}$ to a unique key and can not account for the quantization errors that inevitably happen due to the noise. To circumvent these quantization artifacts, resulting from the hard assignment in ~\cite{drost2010model}, we quantize $\mathbf{f}_{ri}$ to $K$ different bins ($K>1$), activating different codebook entries as in ISM. This soft quantization results in possibly matching buckets $\mathbf{\bar{F}}_{ri}=\{\mathbf{\bar{f}}_{1}..\mathbf{\bar{f}}_{K}\}$. $\mathbf{\bar{F}}_{ri}$ indexes the buckets of $\mathbf{H}$, with weights $p(\mathbf{\bar{f}}_{k}|\mathbf{f}_{ri})$. For each matching bucket, we collect the valid interpretations $p(m,\alpha|\mathbf{\bar{f}})$, inversely proportional to the size of the bucket $N_b$, denoting the probabilities of particular pose configuration, given the quantized feature. Formally: 
\vspace{-3pt}
\begin{align}
p(\mathbf{m},\alpha |\mathbf{s}_r, \mathbf{s}_i) &= p(\mathbf{m},\alpha|\mathbf{f})\\
&= \sum\limits_k p(\mathbf{m},\alpha | \mathbf{\bar{f}}_k) p(\mathbf{\bar{f}}_k | \mathbf{f})
\end{align}
\vspace{-2pt}
with $p(\mathbf{\bar{f}}_k | \mathbf{f})=\frac{1}{K}$ and $p(m,\alpha|\mathbf{\bar{f}})=\frac{1}{N_b}$ being uniformly distributed. This probability is actually the prior on the PPF of the particular object and can be computed differently accounting for the nature of the object geometry using a suited distribution. 
At this point, the gathered pair representations for a particular scene reference point are sufficient to recover for the object pose. However, due to outliers, some of these matches will be erroneous. Therefore, a 2D voting scheme is employed, locally for each scene reference point $\mathbf{s}_r$. The voting space is composed of the alignment of the LCF $\alpha$ as well as the model point correspondence $\mathbf{m}$:
\vspace{-1pt}
\begin{equation}
	V(\mathbf{m}, \alpha) = \sum\limits_i p(\mathbf{m},\alpha |\mathbf{s}_r, \mathbf{s}_i)
\end{equation}
\vspace{-1pt}
For each $\mathbf{s}_r$, there is a voting space $V_r(\mathbf{m},\alpha)$, from which the best alignment is extracted as:
\vspace{-1pt}
\begin{equation}
	(\mathbf{m}^*_r, \alpha^*_r) = \argmax_{\mathbf{m},\alpha} V_r(\mathbf{m},\alpha)
\end{equation}
\vspace{-1pt}
Each such $(\mathbf{m}^*_r, \alpha^*_r)$ corresponds to a pose hypothesis. This is similar to performing Generalized Hough Transform (GHT) on reference point level locally and is the reason why we attribute the name Local ISM to our method. 
	After all pose hypotheses are extracted, as the maxima in the local spaces, the poses are clustered together to assemble the final consensus, further boosting the final confidence.

\paragraph{Hypotheses Verification and Rejection} Devised matching theoretically generates a pose hypothesis for each scene reference point, which is assumed to be found on the model. There are typically $\sim 400-1000$ such points, reducing to ~50 poses after the clustering, where the close-by poses are grouped together and averaged. Still, as many hypotheses as the number of clusters remain to be verified and the best pose is expected to be refined. In our problem of instance reconstruction it is critical that no false positive pose hypotheses survives. For this reason we introduce a rigorous hypothesis verification scheme. The effective verification requires fine registration, while efficient registration requires as few poses as possible. This creates a chicken and egg problem. We address this issue via a multi-level registration approach. In the first stage, sparsely sampled scan points are finely registered to the model using the efficient LM-ICP \cite{fitzgibbon2003robust} variant of Iterative Closest Point (ICP) registration \cite{besl1992method}. We also build a 3D distance transform for fast nearest neighbor access. Our sparse LM-ICP requires only 1ms per hypothesis, allowing us to verify all the hypotheses. We define the hypothesis score to be:
\begin{equation}
\Xi (\bm{\theta}_i ) = \frac{1}{N_M} \sum\limits_j^{N_M} 
\begin{cases}
1, &  \lVert \bm{\theta}^{-1}_i \circ \mathbf{m}_j - \mathbf{s}_k(j) ) \rVert<\tau_\theta \\ 
0, & otherwise
\end{cases}
\end{equation}
where $\bm{\theta}_i$ is the pose hypothesis and $\mathbf{s}_k(j)$ the closest sampled scene point to transformed model point $\bm{\theta}_i^{-1}\circ \mathbf{m}_j$. Intuitively, this score reflects the percentage of visible model points.
The surviving poses are then sorted, taken to the next level and densely refined. This coarse to fine scheme is repeated for 3 levels of the pyramid. Finally, a dense registration is performed to accurately obtain the final pose.

Until this stage the surface normals are excluded from the fine registration process. We do this intentionally, to use them as a verification tool. Following registration, we check the surface consistency between the scene and the model. To do so for each scene point, the surface normal of the closest model point is retrieved. A scan is only accepted if a majority of the normals agree with the model. While this procedure can result in potentially good detections being removed (due to scene deviations), it does not allow false positives to survive as shown in Sec. \ref{sec:results}.

\subsection{Computing Pose Graph $A$ and Live Feedback}
Any global optimization algorithm requires an adjacency graph $G=(V,E)$, which encodes the existence of overlap between camera views. The nodes of this sparse graph contain the cameras $V=\{C_1..C_N\}$, whereas an edge $E_{ij}$ is only created between nodes $(C_i, C_j)$ if they share significant overlap. An absolute pose $\mathbf{T}_i$ is associated to each node and a relative pose $\mathbf{T}_{ij}$ is to each edge. Traditionally, this requires pair-wise overlap computation between all cameras. While a naive approach would involve linking the cameras, whose centers are found to be close, this is by no means a guarantee for shared overlap. Therefore, we present a more accurate approach, without sacrificing efficiency, thanks to the availability of the CAD model. 

Consider the voxel grid index $\mathbf{D}$ of model $\mathbf{M}$ as in Fig. \ref{fig:pose_graph}{\textcolor{red}{(a)}}. Each segmented scene point $\mathbf{s}_{i}' \in \mathbf{T}^{-1}_i \circ \mathbf{S}_i$ is mapped to a voxel $D_k$, which stores a set of cameras $\{C_i\}$ observing it. Whenever the point $\mathbf{m}_k$ belonging to the voxel $D_k$ is visible in the camera $C_j$, this camera is added to the list of cameras seeing that model point. Each list stores unique camera indices. From that, we compute the histogram of pairwise overlaps (HPO) as shown in Fig. \ref{fig:pose_graph}{\textcolor{red}{(b)}}.

\insertimageC{1}{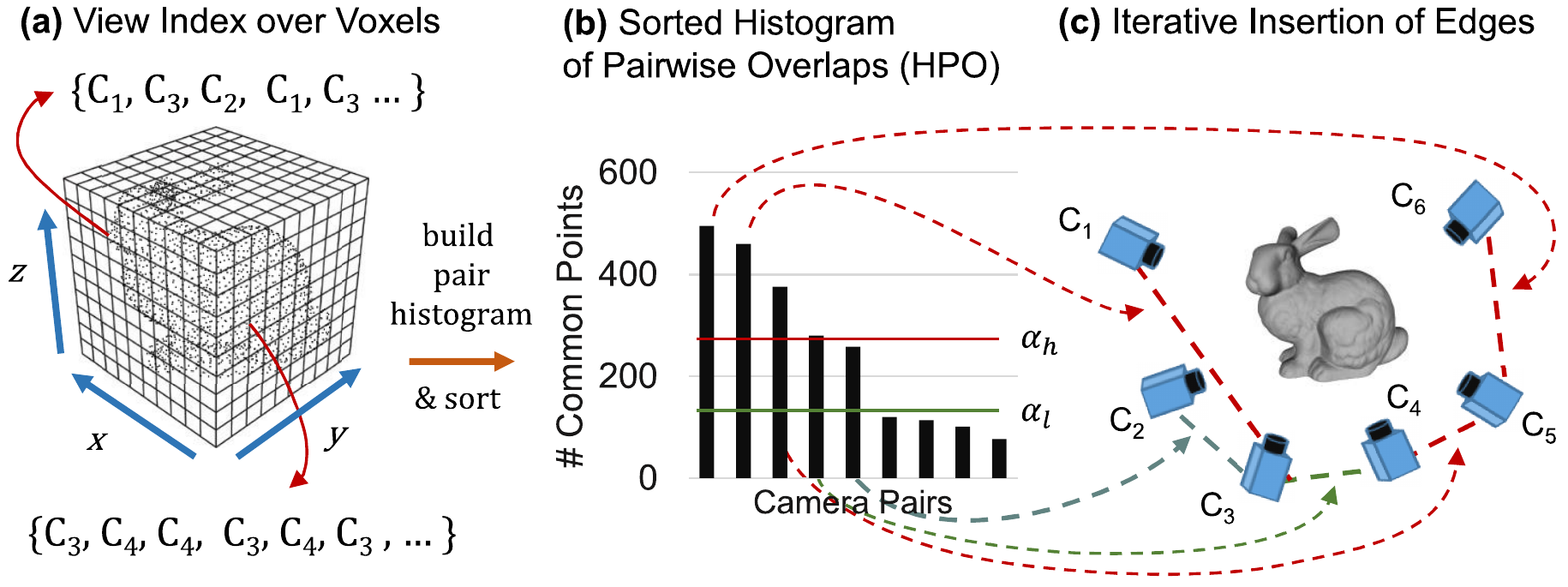}{\small{Pose graph computation. \textit{See text for details.}}}{fig:pose_graph}{t!}
While all the possible edges are now generated (as the bins in HPO), it is not recommended to use all these in the multiview alignment i.e. the overlap might be little, causing a negative impact. Instead, we adopt an iterative algorithm, similar to hysteresis thresholding. First, HPO is sorted with decreasing overlap (Fig. \ref{fig:pose_graph}{\textcolor{red}{(c)}}). Next, two thresholds $\alpha_l$ and $\alpha_h$ are defined. All pairs with overlap less than $\alpha_l$ are discarded. All cameras with overlap larger than $\alpha_h$ are immediately linked and edges are constructed in the graph. If, at this stage, the graph is not connected, we start inserting edges from the remaining bins of HPO into $A$ until either the connectivity or the threshold $\alpha_l$ is reached. This is illustrated in Figures \ref{fig:pose_graph}{\textcolor{red}{(c)}} and \ref{fig:pose_graph}{\textcolor{red}{(d)}}. If the final graph is still not connected, we use the largest connected sub-graph, to ensure optimize-ability. For efficient online update, a modified union-find data structure is used to store the graph and dynamically insert edges when new views are encountered. Unlike quadratic complexity of the standard pose graph creation methods, ours has linear complexity.
\vspace{-10pt}
\paragraph{Live Feedback} Due to the connected-ness of pose graph, our method is able to keep track of the overlap between all the point clouds, at all times, informing the user whenever graph disconnects or overlap is small. The complement of the already reconstructed part reveals the unscanned region, which is also fed back to the operator. Incoming scans directly propagate and form links in the pose graph, allowing online response to the user's actions. 
\insertimageC{1}{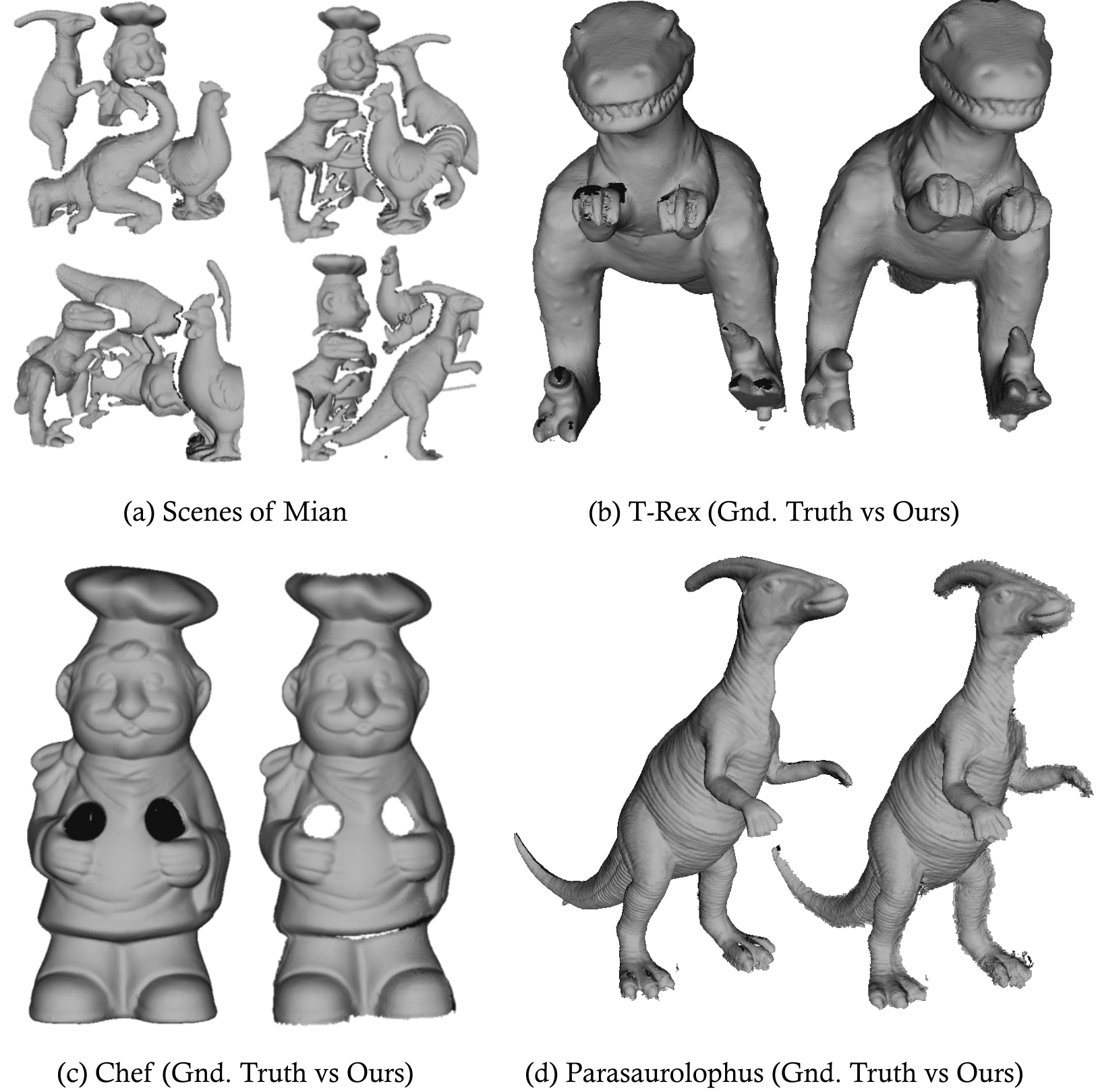}{\small{Results on Mian Dataset. \textbf{(a)} Subset of scenes from the dataset. \textbf{(b, c, d)} The ground truth models (left) and our reconstruction (right) for three objects.}}{fig:mian}{t!}
\section{Experimental Evaluation}
\label{sec:results}
We evaluate our method against a set of real datasets acquired by laser scanners and structured light sensors. The CAD models we work with might contain uneven distribution of vertices or inner geometry. We always eliminate the inner structure by thresholding the ambient occlusion values \cite{miller1994efficient} before the models are re-meshed \cite{levy2010p}. At detection time, a relative model and scene sampling distance of $d=\tau diam(\bm{M})$ is used, where $0.05 \geq \tau \geq 0.025$ depending on the object. We also adjust another threshold on the distance to consider a scene point to be on the model based on the sensor quality. For accurate scanners we use $1.5mm$, while for less accurate ones $0.5cm$. This does not affect the segmentation, but the hypothesis verification.


\paragraph{Mian Dataset}
We first compare the reconstruction quality on Mian Dataset \cite{mian2006three}. This dataset includes 50 laser scanned point clouds of 4 complete 3D objects, with varying occlusion and clutter. The objects change locations from scan to scan, creating dynamic scenes. The clutter and background also varies as the objects appear together with other different ones in each scene. We quantify this dynamic clutter by relating it to the provided occlusion values:
\begin{align}
\text{Clutter} &= 1-\frac{\text{(Model Surface Area)*(1-\text{Occlusion})}}{\text{(Scene Surface Area)}}
\end{align}
and provide it in Table \ref{tab:mian} for each object. The models present in the scenes are provided by \cite{mian2006three} to act as ground truth. We do not perform any prior operation to the scenes such as segmentation or post-processing except meshing via SSD \cite{taubin2012smooth}. For Parasaurolophus and Chicken objects, the pose graph becomes disconnected and therefore, we optimize individually the two sub-components and record the mean. We also report the number of scans in which the model is detected and verified. Not every model is visible in every scan. In the end, our mean accuracy is well below a millimeter, where the used sensor, Minolta Vivid 910 scanner, reports an ideal accuracy of $\sim0.5$mm. We are also not aware of any other works, reporting reconstruction results on such datasets. Fig. \ref{fig:mian} visualizes our outcome, and Table \ref{tab:mian} shows our reconstruction accuracy both prior to and after the optimization. While our error is quantitatively small, the qualitative comparison also yields a pleasing result, sometimes being superior even to the original model.

\begin{table}[t!]
\captionsetup[table]{aboveskip=0pt}
\captionsetup[table]{belowskip=0pt}
  \centering
  \setlength{\tabcolsep}{5pt}
  \setlength\extrarowheight{-0.75pt}
  \caption{\small{Reconstruction results on Mian Dataset (in mm). Each object is compared to the model provided by \cite{mian2006three} using \cite{cloudcompare}.}}
  \small{
    \begin{tabular}{lcccc}
         Model & w/o Opt. & with Opt. & $\#$ Scans & Clutter\\
    \toprule
    \textbf{Chef}  & 2.90 $\pm$ 2.40   & 1.07  $\pm$ 0.65 & 22 & 0.58$\pm$0.11 \\
    \textbf{Chicken} & 1.71 $\pm$ 1.60  & 0.33 $\pm$ 0.24 & 29 & 0.61$\pm$0.12 \\
    \textbf{Para.} & 2.52  $\pm$ 2.00  & 0.41 $\pm$ 0.30 & 12 & 0.24$\pm$0.20\\
    \textbf{T-rex} &     2.36 $\pm$ 2.08     & 0.88  $\pm$ 0.62 & 27 &0.14$\pm$0.22\\
    \toprule
    \end{tabular}%
    }
  \label{tab:mian}%
\end{table}%

\begin{figure}[htbp]
\begin{center}
\subfigbottomskip =0cm
\subfigure[PR-curve without Verification.]{
\includegraphics[width=0.45\columnwidth]{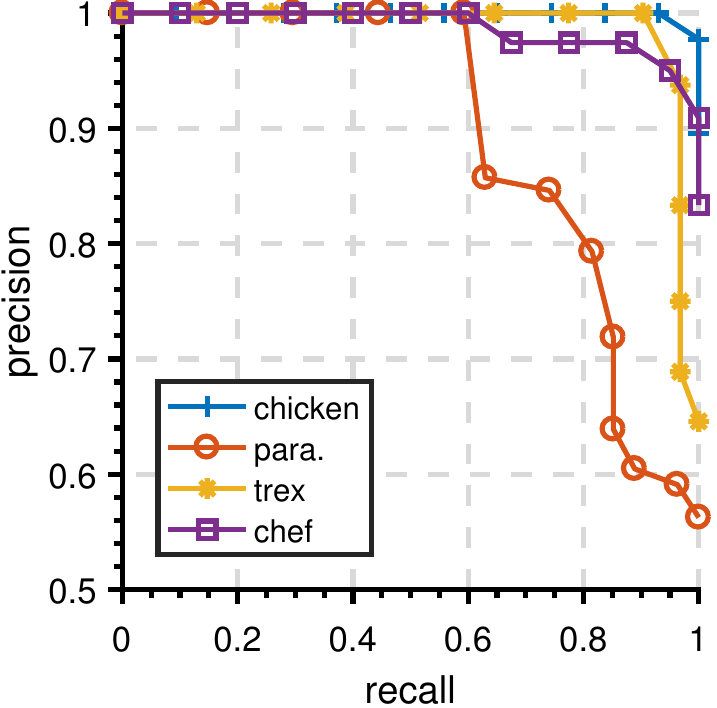}
\label{fig:recall_no_verif}}
\subfigure[PR-curve with verification.]{
\includegraphics[width=0.45\columnwidth]{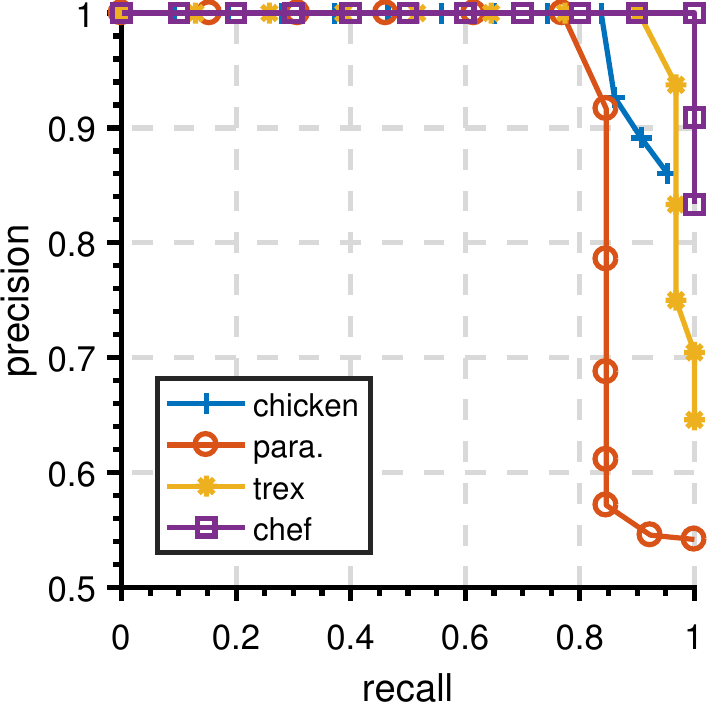}
\label{fig:recall_verif}}
\subfigure[Error Histo. in Toy Objects]{
\label{fig:histograms}
\includegraphics[width=0.485\columnwidth]{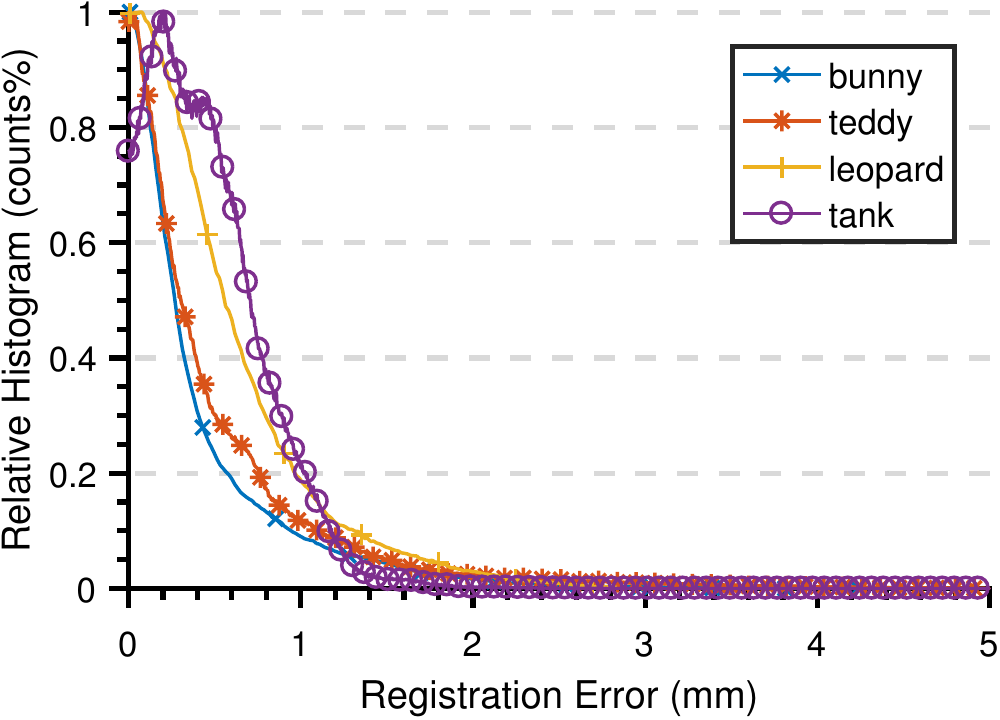}}
\subfigure[Sensitivity to CAD prior]{
\label{fig:decimate_exp}
\includegraphics[width=0.44\columnwidth]{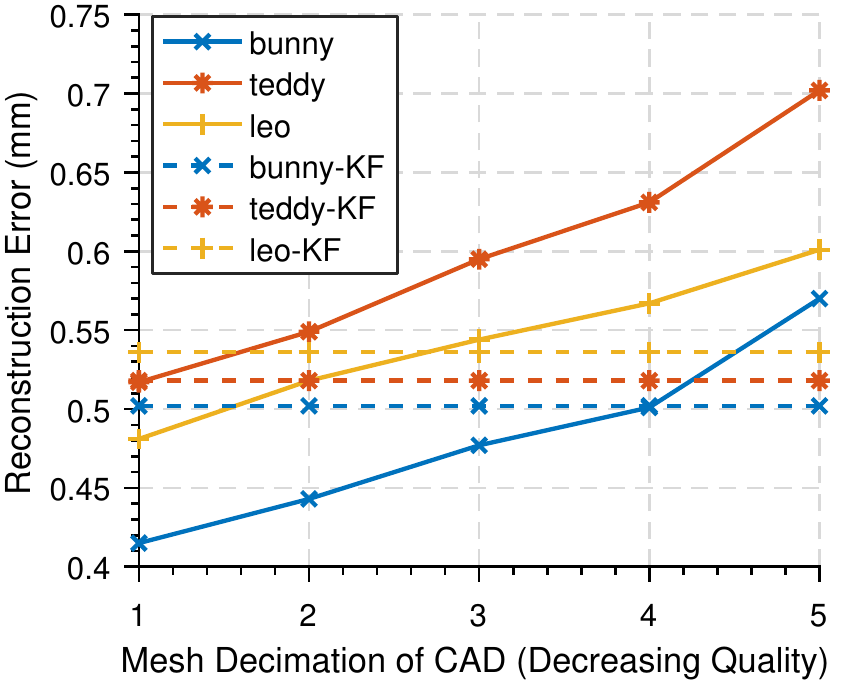}}
\end{center}
\subfigbottomskip =-1cm
   \caption{\small{Performance of LISM and verification on Mian dataset.}}
\end{figure}

The detection performance of a basic variant of our method has already been proven to be robust on this dataset \cite{drost2010model}. Figures \ref{fig:recall_no_verif} and  \ref{fig:recall_verif} provide PR-curves for LISM and the hypothesis verification. Note that although LISM already performs well, our verification clearly improves the distinction between a match vs false positive. Using a simple threshold, we could obtain $100\%$ precision without sacrificing the recall. Thus, our score threshold, combined with the normal consistency check manages to reject all false hypotheses, at the expense of rejecting a small amount of TP.

\vspace{-5pt}
\paragraph{Toy Objects Dataset}
Since our objective is to assess the fidelity of the reconstruction to the CAD model, we opt to use the objects from the 3D printed dataset \cite{slavcheva2016eccv}: \textit{Leopard}, \textit{Teddy} and \textit{Bunny} and \textit{Tank} (See Fig. \ref{fig:toy_vis}). The diameters of objects vary in the range of $15-30cm$. The print accuracy is up to $50\mu$ (micro-meters), well sufficient for consideration as ground truth. To capture the real scenes, a home-brew, high accuracy phase-shift sensor, delivering $<$0.4mm point accuracy is chosen. We sample up to 10 scans per object, taken out of a 100 frame sequence. To disrupt the acquisition order, we randomly shuffle this subset and apply our reconstruction algorithm. Next, we compute the CAD-to-reconstruction distances in CloudCompare~\cite{cloudcompare}. We do not explicitly register our reconstruction to CAD model, because we already end up on model coordinate frame (Having the result in the CAD space is a side benefit of our approach). Moreover, we use the original 100-frame, ordered sequence as an input to standard reconstruction pipelines such as Kinect Fusion \cite{Rusu_ICRA2011_PCL}, Kehl et. al. \cite{kehl2014bmvc} (also uses color) and Slavcheva et. al. \cite{slavcheva2016eccv} all of which require a temporally ordered set of frames, with a large inter-frame overlap. All of these algorithms take depth image as input, whereas ours uses the unstructured 3D data and the model. 

\begin{table}[b!]
\vspace{-1pt}
\captionof{table}{\small{Reconstruction errors on toy objects dataset (mm).}}
\label{tab:toy}
\setlength{\tabcolsep}{3pt}
\resizebox{\columnwidth}{!}{
\begin{tabular}{lcccc}
     & \textbf{Leo} & \textbf{Teddy} & \textbf{Bunny} & \textbf{Tank}\\
    \midrule
    KinFU& 1.785$\pm$1.299 & 0.998$\pm$0.807 & 0.664$\pm$0.654 & 1.390 $\pm$ 1.315 \\
    Kehl& 1.018$\pm$1.378 & 1.028$\pm$0.892 & 0.838$\pm$0.860 & 1.573$\pm$2.250\\
    Sdf2Sdf& 0.652$\pm$0.614 & 0.910$\pm$0.584 & 0.541$\pm$0.436 & 0.466 $\pm$ 0.416\\
    Ours & \textbf{0.481}$\pm$\textbf{0.519} & \textbf{0.517}$\pm$\textbf{0.572} & \textbf{0.415}$\pm$\textbf{0.501} & \textbf{0.451}$\pm$\textbf{0.322} \\
    Ours-KF. & 0.536$\pm$0.411& 0.519$\pm$0.582 & 0.502$\pm$0.529 & 0.468$\pm$0.474 \\
    Ours-CO. & 0.651$\pm$0.628& 0.544$\pm$0.601 & 0.698$\pm$0.506 & 0.475$\pm$0.433 \\
    \bottomrule
    \end{tabular}}%
\end{table}
Our results on this dataset are shown in Tab. \ref{tab:toy} (\textit{Ours}) when original CAD is used. We also report the results when KinectFusion (KF) prior is used (\textit{Ours-KF}). The individual error distribution of the objects are shown in Fig. \ref{fig:histograms}. Because our method does not suffer from drift and computes absolute poses all the time, although we use 10 times less scans, we are still 2-4 times more accurate then conventional methods. This also shows that our method could retain the good accuracy of the sensor.

\insertimageC{1}{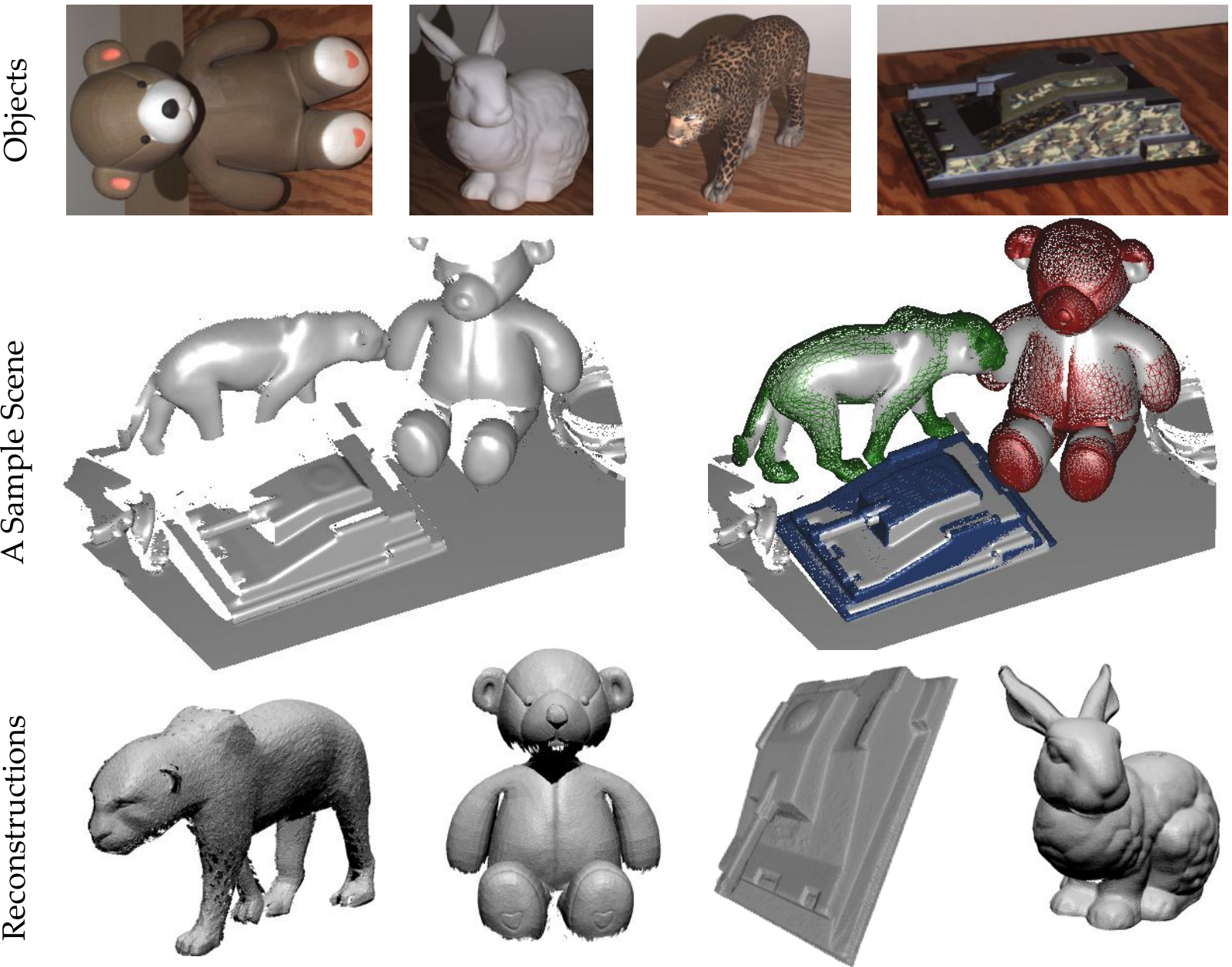}{\small{Qualitative results on toy objects. First row: Real images of objects; Second row: A sample scene and detections visualized; Third Row: Our results.}}{fig:toy_vis}{t!}
Next, we augment this dataset with further scenes of the same objects, such that clutter and occlusions are present. Some shots are shown in Fig. \ref{fig:toy_vis} (mid-row). Our reconstruction accuracy (\textit{Ours-CO}) is shown in Tab. \ref{tab:toy} for different objects. These results are still better than or close to Sdf2Sdf \cite{slavcheva2016eccv}. Due to inclusion of some outliers, our results get slightly worse than the one in no clutter, yet they are still acceptable. However, none of the other approaches can run on this new set due to the existence of significant outliers.

In a further experiment, we gradually decimate the toy models down to a mesh of $\approx$500 vertices. We exclude tank as the decimation has little effect on the planarities. As shown in Fig. \ref{fig:decimate_exp}, even though the CAD prior gets very crude, we are still able to achieve a reasonable reconstruction, as long as the CAD model is still detectable in the scenes. Note that, results of KF prior is plotted in dashes as it is also a form of rough mesh approximation. Furthermore, Fig. \ref{fig:tank_obj} visually compares our reconstruction to the state of the art on the tank object. Because we do not use smoothing voxel representations (such as SDF), our method is much better at preserving sharp features at the model edges.
\insertimageC{1}{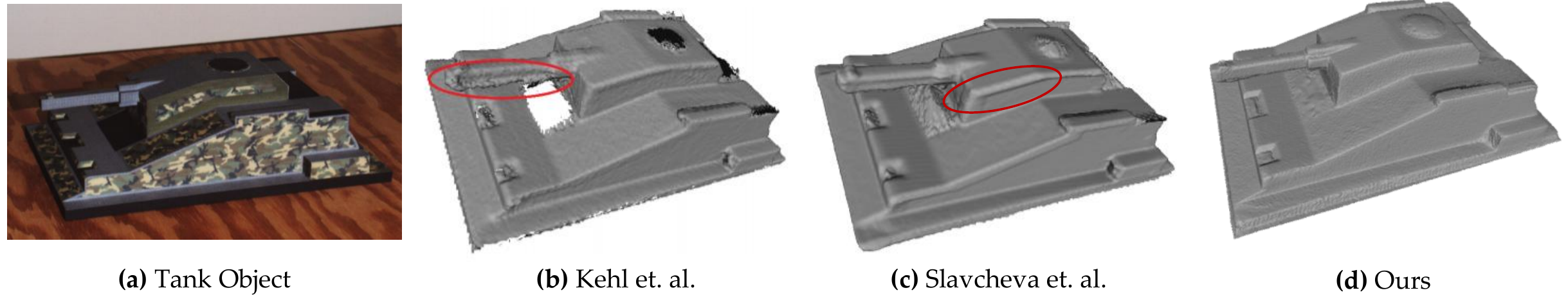}{\small{Visual comparisons on Tank object. Note the ability of our method in preserving sharp features.}}{fig:tank_obj}{htbp}

\insertimageStar{1}{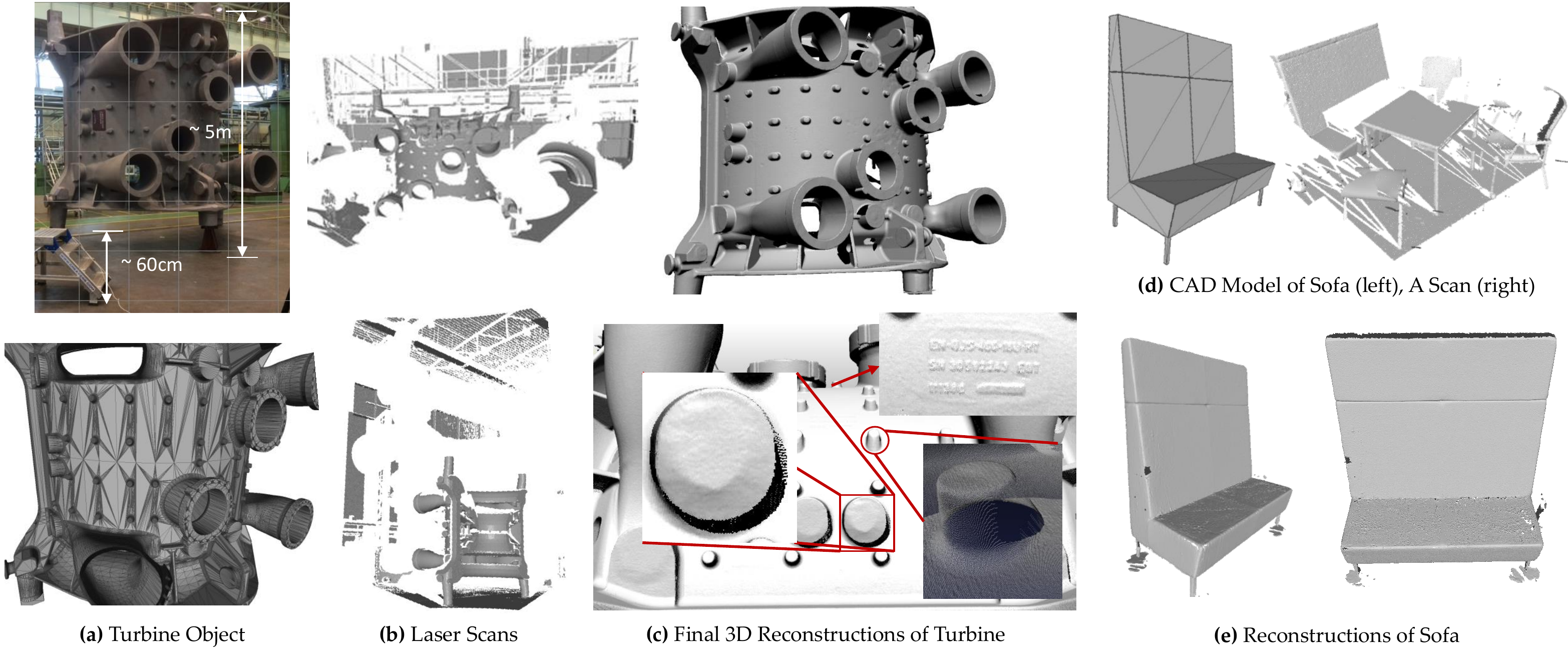}{\small{The reconstruction of Turbine(a) in captured cluttered scans(b) is presented in (c). Results in Sofa are shown in (d,e).}}{fig:turbine_rec}{!t!}
\begin{table*}[!htb]
  \centering
  \setlength\extrarowheight{-0.75pt}
  \caption{\small{Object Information, Average reconstruction errors w.r.t. Photogrammetry (in mm) and Timings.}}
  \resizebox{\linewidth}{!}{
    \begin{tabular}{cccccccccccc}
    \toprule
          Object&Scanner&Scan Res.&Obj. Size&No Scans&No PG Images&PG vs CAD&Surphaser&Our Accuracy&Detect&Verify&Refine \\
    \midrule
    Ventil &Surph. & $0.3$ mm &$8m^3$ & 8 & 180 &1.3cm& 3.6$\pm$3.3 & \textbf{2.2$\pm$0.4}&3.10s&0.27s&112.94s\\
    Turbine &Surph. & $0.4$mm & $125m^3$&10&180& 3.4cm&-& \textbf{2.5$\pm$1.3}&3.72s&0.54s&126.13s\\
    Sofa &Str.Light& $1$mm & $1.7m^3$&6&68&0.85mm&-& \textbf{1.4$\pm$1.2}&1.44s&0.31s&68.82s\\
    \bottomrule
    \end{tabular}}%
  \label{tab:photo}%
\end{table*}%
\paragraph{Dataset of Large Objects} 
Finally, we apply our pipeline to quality inspection of real gas turbine casings and large objects. In this real scenario, CAD models come directly from the manufacturer. Due to space constraints, we summarize the data modality in Tab.\ref{tab:photo}. The manufactured parts deviate significantly from the ideal model due to manufacturing and we scan them in the production environment within clutter and occlusions. With such large objects and little resemblance of the CAD prior, obtaining ground truth becomes a challenging task. Thus, we use a photogrammetry (PG) system~\cite{birdal2016x,birdal2016online} to collect a sparse set of scene points, by attaching markers on the objects. We capture many images  (see Tab.\ref{tab:photo}) from different angles and run Linearis 3D software for bundle adjustment to obtain sparse ground truth. For Ventil object, we also use Surphaser software for reconstruction using external markers. Both Surphaser and our outputs are compared to the PG data in Tab. \ref{tab:photo}. The mean errors are obtained by CloudCompare \cite{cloudcompare}. We also provide running times of the individual stages. As seen, our accuracy outperforms an industry standard solution, Surphaser Software, by a margin of $38\%$ on Ventil object. The performance in objects of varying sizes indicate that our reconstruction method is applicable from small to large scale while maintaining repeatability. Fig. \ref{fig:turbine_rec} presents further qualitative results on our reconstruction of the Turbine and Sofa objects. Please consult the supplementary material for more evaluations.
\vspace{-5pt}
\paragraph{Limitations} 
Due to the nature of PPF matching, our approach requires objects with rich geometry. Symmetric objects are also problematic due to ambiguity in pose estimation. Last but not least, currently, there is no mechanism to handle mis-detections. Yet, mis-detections are hardly a problem when the score threshold is reasonably high. This way, we detect in less scenes but avoid mistakes.

\section{Conclusions}
We proposed \textit{reconstruction-via-detection} framework, as an alternative perspective to robust 3D instance reconstruction from unconstrained point cloud scans. Our framework integrates probabilistic object detection, hypothesis verification, pose graph construction and multi-view optimization. Such a scheme allowed us to deal with problems of dynamic clutter, occlusion and object segmentation. Moreover, the computational cost is reduced, due to model-to-scan alignment. To the best of our knowledge, this is the first method, capable of reconstructing instances within clutter and occlusions, without explicit segmentation.

As a future direction, we like to take care of confusions stemming from rotational symmetries by optimizing over possible global alignments of the scans. We also plan to extend our method to robotics via next-best view prediction.

\clearpage
{\small
\bibliographystyle{ieee}
\bibliography{egbib}
}

\end{document}